\documentclass[runningheads]{llncs}
\usepackage{graphicx}
\usepackage{cite}
\usepackage{xcolor}
\usepackage{tikz}
\usepackage{subcaption}
\usepackage{multirow}

\usepackage[final]{microtype}
\usepackage[british]{babel}
\usepackage{hyphenat}
\usepackage{istgame}
\usepackage{booktabs}
\usepackage{amsmath,amsfonts}
\usepackage[absolute,showboxes]{textpos}
\TPMargin{8pt}

\usepackage{hyperref}

\hypersetup{
  pdffitwindow=false,     
  pdfstartpage={1},
  pdfstartview={Fit},    
  pdfnewwindow=true,      
  colorlinks=true,       
  allcolors=blue,
  breaklinks=true
}
\begin{document}

\begin{textblock}{0.7}(0.16,0.88) 
    \noindent
    \footnotesize
    Derks I.P., de Waal A. (2020) A Taxonomy of Explainable Bayesian Networks. In: Gerber A. (eds) Artificial Intelligence Research. SACAIR 2021. Communications in Computer and Information Science, vol 1342. Springer, Cham.\\
    \textbf{\url{https://doi.org/10.1007/978-3-030-66151-9_14}}
\end{textblock}

\title{A Taxonomy of Explainable Bayesian Networks}
%
%
\author{Iena Petronella Derks\inst{1}\orcidID{0000-0002-7070-5036} \and
Alta de Waal\inst{1,2}\orcidID{0000-0001-8121-6249}}
\authorrunning{IP Derks, A de Waal}
%
\institute{Department of Statistics, University of Pretoria \and Center for Artificial Intelligence Research (CAIR)}
\maketitle              
\begin{abstract}
Artificial Intelligence (AI), and in particular, the explainability thereof, has gained phenomenal attention over the last few years. Whilst we usually do not question the decision-making process of these systems in situations where only the outcome is of interest, we do however pay close attention when these systems are applied in areas where the decisions directly influence the lives of humans. It is especially noisy and uncertain observations close to the decision boundary which results in predictions which cannot necessarily be explained that may foster mistrust among end-users. This drew attention to AI methods for which the outcomes can be explained. Bayesian networks are probabilistic graphical models that can be used as a tool to manage uncertainty. The probabilistic framework of a Bayesian network allows for explainability in the model, reasoning and evidence. The use of these methods is mostly ad hoc and not as well organised as explainability methods in the wider AI research field. As such, we introduce a taxonomy of explainability in Bayesian networks. We extend the existing categorisation of explainability in the model, reasoning or evidence to include explanation of decisions. The explanations obtained from the explainability methods are illustrated by means of a simple medical diagnostic scenario. The taxonomy introduced in this paper has the potential not only to encourage end-users to efficiently communicate outcomes obtained, but also support their understanding of how and, more importantly, why certain predictions were made.

\end{abstract}

\section{Introduction} 
Advances in technology have contributed to the generation of big data in nearly all fields of science, giving rise to new challenges with respect to explainability of models and techniques used to analyse such data. These models and techniques are often too complex; concealing the knowledge within the machine, hence decreasing the extent of interpretability of results. Subsequently, the lack of explainable models and techniques contribute to mistrust among users in fields of science where interpretability and explainability are indispensable. 

To elucidate the need for explainable models, consider the following three scenarios. Firstly, suppose a medical diagnosis system is used to determine whether a tumour sample is malignant or benign. Here, the medical practitioner must be able to understand how and why the system reached the decision, and, if necessary, inspect whether the decision is supported by medical knowledge \cite{brito2019explainable}. Next, consider self-driving cars. In this context, the self-driving car must be able to process information faster than a human, such that accidents and fatalities can be avoided \cite{lawless2019artificial}. Suppose a self-driving car is involved in an accident, then the system must be able to explain that in order to avoid hitting a pedestrian, the only option was to swerve out of the way and, by coincidence, into another vehicle. Lastly, consider an online restaurant review system, where reviews are classified as positive or negative based on the words contained in the review. Here, the classifier simply returns whether a review is positive or negative, without explaining which words contributed to the classification. As such, negative reviews that are expressed in, for example, a sarcastic manner, might be classified as positive, resulting in a restaurant receiving a higher rating and more diners -- who might experience bad service (or even food poisoning) as a result of mislabelled reviews.

Given its relevance in many application areas, the explainability problem has attracted a great deal of attention in recent years, and as such, is an open research area \cite{Lipton2018}. The manifestation of explainable systems in high-risk areas has influenced the development of explainable artificial intelligence (XAI) in the sense of prescriptions or taxonomies of explanation. These include fairness, accountability, transparency and ethicality \cite{Cath2018,Greene2019,Leslie2019}. The foundation of such a system should include these prescriptions such that a level of usable intelligence is reached to not only understand model behaviour \cite{BarredoArrieta2020} but also understand the context of an application task \cite{Holzinger2017b}. Bayesian networks (BNs) -- which lie at the intersection of AI, machine learning, and statistics -- are probabilistic graphical models that can be used as a tool to manage uncertainty. These graphical models allow the user to reason about uncertainty in the problem domain by updating ones beliefs, whether this reasoning occurs from cause to effect, or from effect to cause. Reasoning in Bayesian networks is often referred to as what-if questions. The flexibility of a Bayesian network allows for these questions to be predictive, diagnostic and inter-causal. Some what-if questions might be intuitive to formulate, but this is not always the case especially on a diagnostic and inter-causal level. This might result in sub-optimal use of explainability in BNs - especially on an end-user level. Apart from well-established reasoning methods, the probabilistic framework of a Bayesian network also allows for explainability in evidence. These include most probable explanation and most relevant explanation. To extend on the existing explainability methods, we propose an additional approach which considers explanations concerned with the decision-base.

In this paper, we research the current state of explainable models in AI and machine learning tasks, where the domain of interest is BNs. In the current research, explanation is often done by principled approaches to finding explanations for models, reasoning, and evidence. Using this, we are able to formulate a taxonomy of explainable BNs. We extend this taxonomy to include explanation of decisions. This taxonomy will provide end-users with a set of tools to better understand predictions made by BNs and will therefore encourage efficient communication between end-users.  
The paper is structured as follows. We first investigate the community and scope of explainability methods in Section \ref{related}. Thereafter, we introduce explanation in BNs, which includes the formulation of principled approaches, the theoretical properties associated therewith and a hands-on medical diagnosis example. Section \ref{xbn_workflow} presents our newly formulated taxonomy of explainable BNs. The final section concludes the paper and includes a short discussion of future work.

\section{Related Work}\label{related} 
In application areas where erroneous decisions have a direct impact on livelihood, relying on systems where the predictions cannot be explained may not be an option. Explainability in such systems aids in establishing trust in not only circumstances where the system is used as a primary decision tool, but also cases where the system takes on a supportive role \cite{samek2019towards}. 

Over the past few years, explainability in AI systems has gained immense attention from the research community. This is reflected in the launch of various events and organisations. The Defense Advanced Research Projects Agency (DARPA) launched the Explainable Artificial Intelligence (XAI) initiative in 2016. The XAI programs intention is to encourage the production of AI techniques where emphasis is placed on developing more accurate and precise models, while still maintaining a high level of explainability. Ultimately, XAI systems must be able to explain their rationale and enable understanding \cite{gunning2019darpa}. Conferences, such as the International Joint Conferences on Artificial Intelligence (IJCAI) conducts workshops specifically focusing on XAI \cite{Miller2019}. This topic has also made a noticeable appearance at the Neural Information Processing Systems (NeurIPS) conference, with panel discussions solely concentrating on XAI.

The scope of explainability is inherently linked to the complexity of the model, as well as the goal thereof. Usually, but not necessarily, there is a trade-off between model accuracy and explainability -- the higher the accuracy, the lower the explainability \cite{Xu2019}. For example, decision trees provide a clear explanation but are often less accurate than deep learning models, which are less transparent. It should be mentioned that this trade-off is also connected to the quality of data. AI and machine learning models that are transparent by design, such as linear regression, decision trees and k-nearest neighbours, convey a degree of explainability \cite{BarredoArrieta2020}. However, when AI and machine learning models do not provide clear explanations, separate explainability methods are applied to the model to gain meaningful explanations. Methods of explainability are not limited to the behaviour of the model or decision-making process as a whole, and may be applied to single instances, predictions or decisions \cite{das2020opportunities}. These explanations can be in the form of visualisations or natural language \cite{goebel2018explainable}. Some of the existing explainability methods are layer-wise relevance propagation (LRP), which are often used in deep neural networks where the prediction made by the network is propagated back into the neural network using a set of predefined rules \cite{montavon2019layer}. Another explainability method is local interpretable model-agnostic explanations (LIME). LIME methods can be used to explain prediction instances by attempting to understand the behaviour of the prediction function in the context of the prediction. Here, the user is able to obtain a local explanation for that particular instance. LIME can also be used to obtain explanations for the entire model by generating multiple instances \cite{khedkar2019explainable}. Methods of explainability are also extended to document classifiers, where documents are classified based on predicted likelihood. Here, explanations can be produced based on a search through the text-space of possible word combinations -- starting with a single word and expanding the number of words until an explanation is found \cite{martens2014explaining}.

Uncertainty is present in the majority of AI fields, such as knowledge representation, learning and reasoning \cite{lecue2019role}. Real-world data often contain noisy and uncertain observations close to the decision boundary, which may result in predictions that cannot be explained \cite{de2020uncertainty}. Probabilistic graphical models can be seen as uncertainty management tools as they are able to represent and reason with uncertainty. These probabilistic models are often employed to support decision making in various application fields, including legal and medical applications \cite{timmer2017two}. One such probabilistic model is BNs, which is capable of combining expert knowledge and statistical data, therefore allowing for complex scenarios to be modelled. However, not only are the inner workings of Bayesian networks complicated to most end-users \cite{keppens2016explaining}, the explanation of probabilistic reasoning is challenging and as such results appear to be counter-intuitive or wrong \cite{keppens2019explainable}. Therefore, there exists a demand for explanation in Bayesian networks. 

Explanation methods for Bayesian networks can be divided into three broad approaches. The first approach consists of presenting information contained in the knowledge base and is known as \textit{explanation of the model}. There are two objectives associated with this type of explanation. Firstly, explanation of the model is used to assist application experts in the model-construction phase. Secondly, it is used for instructional purposes to offer knowledge about the domain \cite{Lacave2002}. The objective of the second approach is to justify the conclusion and how it was obtained. This approach is referred to as \textit{explanation of reasoning} \cite{flores2005}. The final approach, \textit{explanation of evidence}, is concerned with the treatment of the variables in the Bayesian network \cite{helldin2009explanation}. In explanation of evidence, also referred to as \textit{abduction}, an explanation is seen as the configuration of a portion of the variables present in the Bayesian network, given evidence. Not included in the aforementioned explanation methods are techniques that describe whether the end-user is ready to make a decision, and if not, what additional information is required to better prepare for decision making. Techniques such as sensitivity analysis \cite{chan2012robustness} and same-decision probability \cite{choi2012same} provide the end-user with insight on decisions. We group these methods into a fourth approach, \textit{explanation of decisions}. For the purpose of this paper, and accordingly, the formulation of the explainable taxonomy, we only consider explanation of reasoning, evidence, and decisions. Explanation of the model is excluded from this taxonomy -- at the time being -- as the intent of the taxonomy is to support understanding of how and why predictions were made and not on the model-construction itself.

\section{Explainable Bayesian networks}\label{methodology}
We adapt the XAI terminology to the scope of BNs by defining the term XBN and thereby referring to explainable BNs. To illustrate XBN in Bayesian networks, consider the Asia network from Lauritzen and Spiegelhalter (1988) \cite{lauritzen1988local} as an example.\\
\textbf{Example statement:} Suppose a patient visits a doctor, complaining about shortness of breath (dyspnoea) (\textbf{D}). The patient is worried he might have lung cancer. The doctor knows that lung cancer is only one of the possible causes for dyspnoea, and other causes include bronchitis (\textbf{B}) and tuberculosis (\textbf{T}). From her training, the doctor knows that smoking increases the probability of lung cancer (\textbf{C}) and bronchitis. Both tuberculosis and lung cancer would result in an abnormal X-ray (\textbf{X}) result. Lastly, a recent visit to Asia might increase the probability of tuberculosis, as the disease is more prevalent there than the patient's country of origin.

\begin{figure}
    \centering
    \includegraphics[scale=0.65]{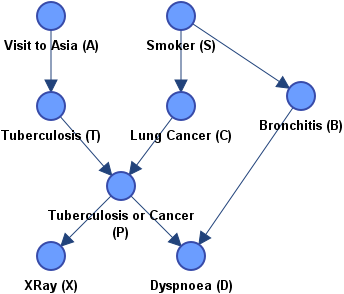}
    \caption{\label{fig:Asia_Layout}Asia Bayesian network}
\end{figure}

From this example statement, the nodes and values are defined and then the graphical structure of the BN is constructed. This is followed by the quantification of the conditional probability tables (CPTs) for each node \cite{korb2010bayesian} \footnote{All BN models are constructed in BayesiaLab (\url{www.bayesialab.com})}. The final BN is illustrated in Figure \ref{fig:Asia_Layout}. Now that the domain and uncertainty are represented in the BN, we will look into how to use the BN. 
Reasoning in BNs takes place once we observe the value of one or more variables and we want to condition on this new information \cite{korb2010bayesian}. It is important to note that this information need not necessarily flow in the direction of the arcs, and therefore, reasoning can occur in the opposite direction of the arcs. 

\subsection{Explanation of reasoning}
\begin{figure}
\begin{subfigure}{0.5\textwidth}
  \centering
  \includegraphics[scale=0.4]{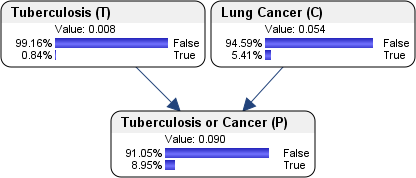}
  \caption{Joint Probability Tables for T, C and P}
  \label{fig:JPT}
\end{subfigure}
\begin{subfigure}{0.5\textwidth}
  \centering
  \includegraphics[scale=0.4]{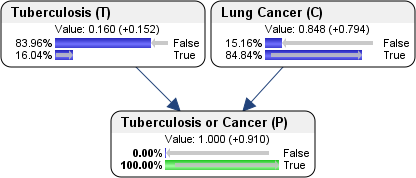}
  \caption{Adding evidence to P}
  \label{fig:JPT_update_P}
\end{subfigure}
\begin{subfigure}{\textwidth}
  \centering
  \includegraphics[scale=0.4]{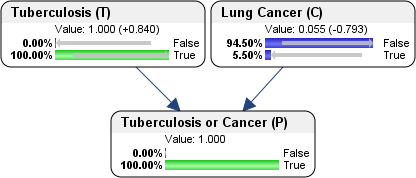}
  \caption{Adding evidence to T}
  \label{fig:JPT_update_T}
\end{subfigure}
\label{fig:fig}
\caption{Belief Updating for T, C and P}
\end{figure}

Suppose during the doctor's appointment, the patient tells the doctor he is a smoker before any symptoms are assessed. As mentioned earlier, the doctor knows smoking increases the probability of the patient having lung cancer and bronchitis. This will, in turn, also influence the expectation of other symptoms, such as the result of the chest X-Ray and shortness of breath. Here, our reasoning is performed from new information about the causes to new beliefs of the effects. This type of reasoning is referred to as \textbf{predictive reasoning} and follows the direction of the arcs in the network. Through predictive reasoning, we are interested in questions concerning \textit{what will happen}. In some cases, predictive reasoning is not of great insight and it is often required to reason from symptoms (effect) to cause, which entails information flow in the opposite direction to the network arcs. For example, bronchitis can be seen as an effect of smoking. Accordingly, we are interested in computing \(P(S|B)\). This is referred to as \textbf{diagnostic reasoning} and is typically used in situations where we want to determine \textit{what went wrong}. The final type of probabilistic reasoning in BNs is \textbf{inter-causal reasoning}, which relates to mutual causes of a common effect -- typically indicated by a v-structure in the network. In other words, inference is performed on the parent nodes of a shared child node. Note that the parent nodes are independent of one another unless the shared child node is observed, a concept known as \textit{d-separation}. From the Asia network, we observe a v-structure between Tuberculosis, Lung Cancer and Tuberculosis or Cancer (see Figure \ref{fig:JPT}). Here, Tuberculosis is independent from Lung cancer. Suppose we observe the patient has either Tuberculosis or Cancer -- indicated by the green (or light grey if viewed in grey-scale) bar in Figure \ref{fig:JPT_update_P} -- then this observation increases the probabilities of the parent nodes, Tuberculosis and Lung Cancer. However, if it is then revealed that the patient does, in fact, have Tuberculosis it, in turn, lowers the probability of a patient having Lung Cancer (see Figure \ref{fig:JPT_update_T}). We can then say Lung Cancer has been \textit{explained away}. It should be noted that the probabilistic reasoning methods discussed above can be used as is, or can be combined to accommodate the problem at hand. 

\subsection{Explanation of evidence}
Sometimes, users of the system find the results of reasoning unclear or questionable. One way to address this is to provide scenarios for which the reasoning outcomes are upheld. A fully specified scenario is easier to understand than a set of reasoning outcomes. Explanation of evidence methods are useful in specifying these scenarios. They are based on the posterior probability and the generalised Bayes factor. Firstly, we focus on methods that aim to find a configuration of variables such that the posterior probability is maximised given the evidence. Here, we consider the Most Probable Explanation (MPE), which is a special case of the Maximum A Posteriori (MAP). The MAP in a BN is a variable configuration which includes a subset of unobserved variables in the explanation set such that the posterior probability -- given evidence -- is maximised. Similarly, if the variable configuration consists of all variables present in the explanation set, we have an MPE solution \cite{helldin2009explanation}. However, in some real-world applications, the variable set often consists of a large number of variables, which may result in over-specified or under-specified explanations obtained from the MPE. In fact, only a few variables may be relevant in explaining the evidence. The next approach finds a single instantiation that maximises the generalised Bayes factor in a trans-dimensional space containing all possible partial instantiations. In other words, this approach aims to obtain an explanation only consisting of the most relevant variables in the BN, given the evidence. This approach is known as the Most Relevant Explanation (MRE) \cite{Yuan2009,yuan2009some,Yuan2011}. 

\subsubsection{Most Probable Explanation}
\begin{figure}
    \centering
    \includegraphics[scale=0.5]{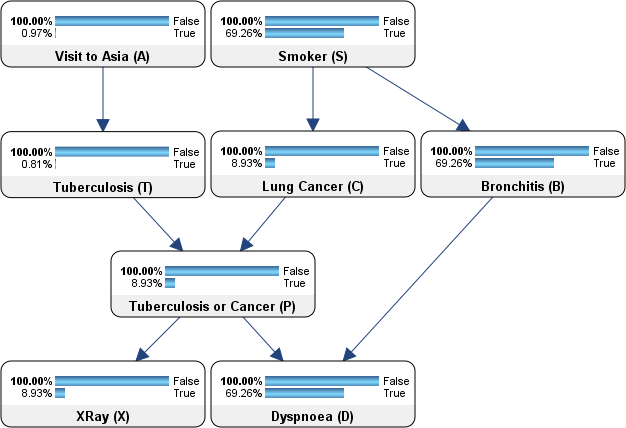}
    \caption{\label{fig:Asia_MPE}Initial MPE for Asia Bayesian network}
\end{figure}

\begin{figure}
    \centering
    \includegraphics[scale=0.5]{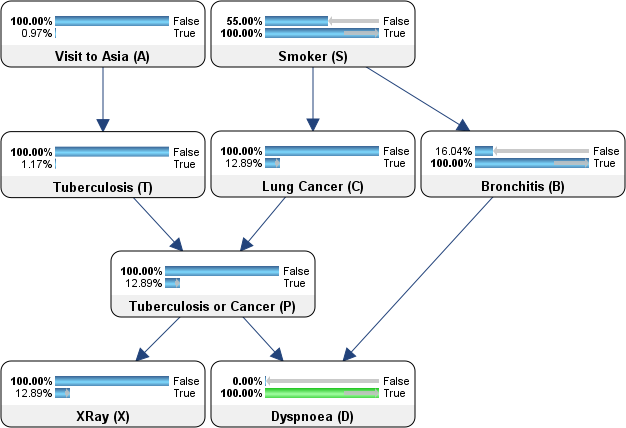}
    \caption{\label{fig:Asia_MPE_updated}Updated MPE for Asia Bayesian network} 
\end{figure}

Let's first consider the MPE method. Recall that the MPE finds the complete instantiation of the target variables -- which are defined to be unobserved -- such that the joint posterior probability is maximised given evidence. Figure \ref{fig:Asia_MPE} shows the scenario (or case) that has the highest joint probability in the Asia network. Note here the probabilities are replaced by the likelihood of the variable state belonging to the most probable scenario, for example, if we look at the two possible states for Bronchitis, we see that `False', i.e., the patient does not have bronchitis, is more probable. Suppose we discover the patient suffers from shortness of breath, we can then set the evidence for Dyspnoea as `True' (illustrated in Figure \ref{fig:Asia_MPE_updated}). By introducing this new evidence, we now observe a slightly different scenario, where it is more probable for the patient to be a smoker and have bronchitis. Notice here that variables that seem irrelevant to the evidence explanation, such as Visit to Asia and XRay, are included in the explanation. This could lead to overspecified hypotheses, especially in larger networks.

\subsubsection{Most Relevant Explanation}
To avoid an overspecified hypotheses, one approach is to trim or prune less relevant variables from the explanation. That is, instead of finding the complete instantiation of the target variables, a partial instantiation of the target variables is found such that the \textit{generalised Bayes factor} is maximised. Let's first consider the explanations obtained from the generalised Bayes factor. Again, suppose the patient suffers from shortness of breath (evidence). We are then interested in finding only those variables that are relevant in explaining why the patient has shortness of breath. Table \ref{tab:GBF} contains the set of explanations obtained from the generalised Bayes factor. For example, the last entry shows that a possible explanation for shortness of breath is a trip to Asia and an abnormal X-ray. Thus including only 2 variables from the remaining 7 variables (excluding Dyspnoea). As mentioned, the MRE is the explanation that maximises the generalised Bayes factor. From Table \ref{tab:GBF} we see that having Bronchitis best explains the shortness of breath. Notice that this explanation does not include Smoking, as opposed to the MPE which included Smoking. Thus, although smoking is a probable cause for shortness of breath, it is not the most relevant cause. An interesting characteristic of the MRE is its ability to capture the \textit{explaining away} phenomenon \cite{Yuan2011}.

\begin{table}[]
    \caption{Explanations of GBF scores for Asia network}
    \centering
    \begin{tabular}{l|l}
        Explanation & Generalised Bayes Factor \\ \hline\hline
        \textbf{(Bronchitis)} & \textbf{6.1391} \\
        (Smoker, Tuberculosis or Cancer) & 1.9818 \\
        (Tuberculosis or Cancer) & 1.9771 \\
        (Lung Cancer, Smoker) & 1.9723 \\
        (Lung Cancer) & 1.9678 \\
        (Smoker, Tuberculosis) & 1.8896 \\
        (Tuberculosis) & 1.8276 \\
        (Smoker, XRay) & 1.7779 \\
        (Smoker) & 1.7322 \\
        (Visit to Asia, XRay) & 1.5635 
    \end{tabular}
    \label{tab:GBF}
\end{table}

\subsection{Explanation of decisions}
Hidden or unobserved variables appear in most application fields, especially in areas where decisions made by the end-user directly influence human lives. For example, when first examining a patient, the health-state of the patient is unknown. In these situations, one would typically ask two questions. The first being \textit{given the available information, are we ready to make a decision?} and secondly, \textit{if we are not yet ready to make a decision, what additional information do we require to make an informed decision?}. To answer these questions, the authors of \cite{choi2012same} propose a threshold-based notion, named \textit{same-decision probability}, which provides the user with a confidence measure that represents the probability that a certain decision will be made, had information pertaining unknown variables been made available. Another possible threshold-based solution to this is sensitivity analysis \cite{van1999sensitivity}. In sensitivity analysis, the assessments for the conditional probabilities in the BN are systematically changed to study the effect on the output produced by the network. The idea is that some conditional probabilities will hardly influence the decisions, while others will have significant impact.

\subsubsection{Same-decision Probability}
\begin{figure}
    \centering
    \includegraphics[scale=0.4]{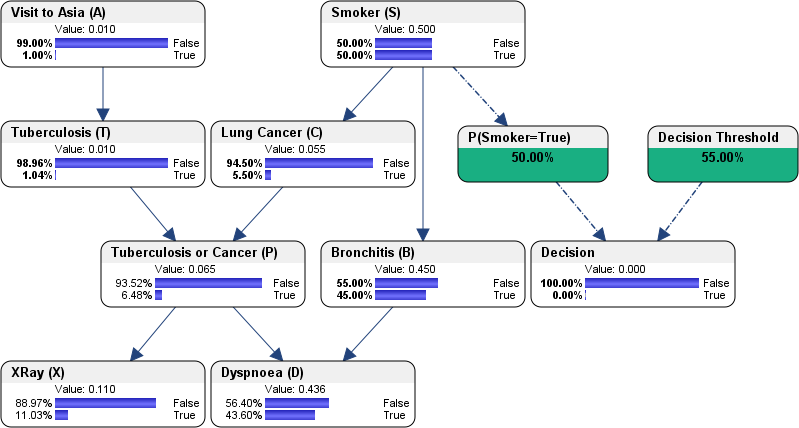}
    \caption{\label{fig:Asia_SDP}Addition of decision node in Asia network}
\end{figure}

\begin{figure}
    \centering
    \includegraphics[scale=0.4]{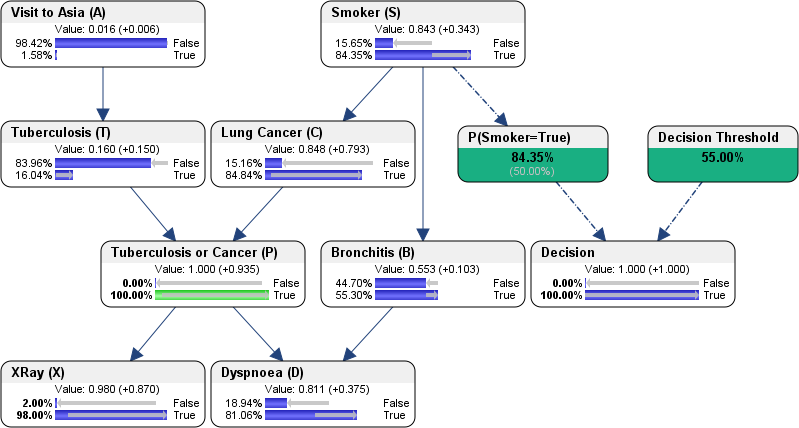}
    \caption{\label{fig:Asia_SDP_update}Updated decision for Asia network}
\end{figure}

Suppose we are interested in making a decision on whether the patient is a smoker (Smoking), which is conditioned on evidence Tuberculosis or Cancer. We can then use the BN such that our decision pertaining to the hypothesis is supported on the basis that the belief in the hypothesis given some evidence exceeds a given threshold. Now, the patient may have access to information that is unknown to us, for example, the patient recently visited Asia and chose not to disclose this information. Therefore, we do not have access to the true state of this variable. The true state knowledge may confirm or contradict our decision based on the probability of smoking given some evidence and the patient visiting Asia. If we now compare this probability with some threshold, we have a degree of confidence in our original decision regarding smoking and the available evidence. Had the patient disclosed his trip to Asia, it is then unlikely that we would have made a different decision. Hence, we can make use of the same-decision probability (SDP). Consider now the BN given in Figure \ref{fig:Asia_SDP}. Notice here the addition of three nodes, \textit{P(Smoker=True)}, \textit{Decision Threshold} and \textit{Decision}. Where \textit{P(Smoker=True}) represents the hypothesis probability and the decision threshold is set to 55\%. Suppose now we update our network such that Tuberculosis or Cancer (P) is True -- to reflect the scenario discussed above. The hypothesis probability then increases from 50.00\% to 84.35\% (see Figure \ref{fig:Asia_SDP_update}). Our decision is confirmed given the threshold since the hypothesis probability now exceeds the given threshold value. From Table \ref{tab:Asia_SDP_confidence}, the SDP before adding evidence for the `True' state is 0.00\%. After adding evidence, the SDP for our decision is 83.88\%, indicating that our decision confidence is 83.88\%\footnote{The SDP scenario was constructed using the decision node functionality in Bayesialab. The decision nodes are indicated as green (or dark grey if viewed in grey-scale).}.

\begin{table}[]
    \centering
    \caption{Decision Confidence for Asia network}
    \begin{tabular}{llllll}
        \multicolumn{1}{l|}{} & States & Minimum & Maximum & Mean & Standard Deviation \\ \hline
        \multicolumn{1}{l|}{\multirow{2}{*}{No evidence}} & False & 100.00\% & 100.00\% & 100.00\% & 0.00\% \\
        \multicolumn{1}{l|}{} & True & 0.00\% & 0.00\% & 0.00\% & 0.00\% \\ \hline
        \multicolumn{1}{l|}{\multirow{2}{*}{Evidence}} & False & 0.00\% & 100.00\% & 16.12\% & 36.77\% \\
        \multicolumn{1}{l|}{} & True & 0.00\% & 100.00\% & 83.88\% & 36.77\%
    \end{tabular}
    \label{tab:Asia_SDP_confidence}
\end{table}

\section{XBN in Action}\label{xbn_workflow}
\begin{figure}
    \centering
    \includegraphics[scale=0.35]{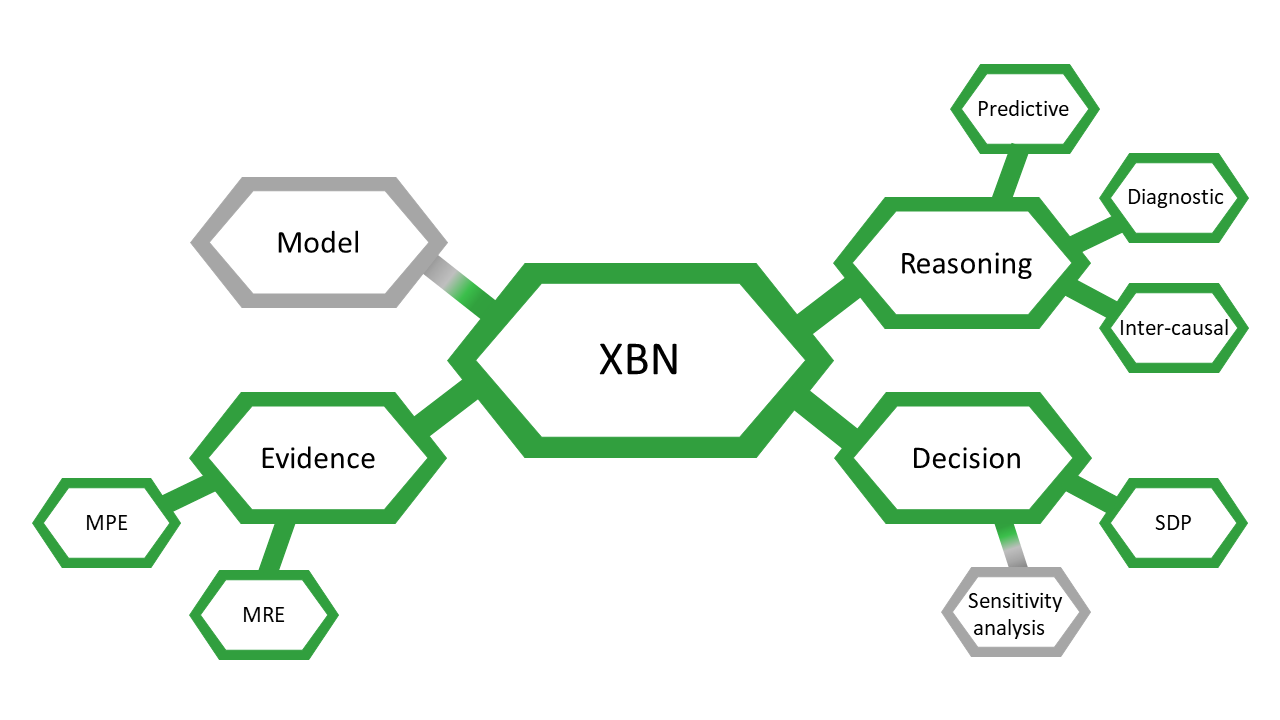}
    \caption{\label{fig:XBN_workflow_chart}A schematic view of XBN}
\end{figure}

The point of XBN is to explain the AI task at hand. In other words, the question the decision-maker seeks to answer, and not the technique in principle. Therefore, we need to be able to freely ask `why' or `what' and from this select a method that would best address the AI task. In Figure \ref{fig:XBN_workflow_chart} we present a taxonomy of XBN. The purpose of this taxonomy is to categorise XBN methods into four phases of BNs: The first phase involves the construction of the BN model. Explanation in the `model' phase is critical when the model is based on expert knowledge. The second phase is reasoning, the third phase evidence, and the fourth decision. Explanation of the model and sensitivity analysis are illustrated in grey as it is out of scope for this paper. Although we define the taxonomy along these phases, we do acknowledge that not all phases are necessarily utilised by the decision-maker. For example, when using BNs to facilitate participatory modelling \cite{duspohl2012review}, the main emphasis is on explaining the model. Or, when using BNs as a classifier, the emphasis is on explaining the decisions. In this section, we present typical questions of interest to the decision-maker in each category of the XBN taxonomy.

\subsection{Reasoning}
Reasoning in the XBN taxonomy is concerned with the justification of a conclusion. Returning to our Asia example, the end-user might ask the following question,
\begin{enumerate}
    \item[] ``\textit{Given the patient recently visited Asia, how likely is an abnormal chest X-Ray?}"
\end{enumerate}
Here, we are concerned with a single outcome: the X-Ray result. On the other hand, the doctor may have knowledge about symptoms presented by the patient and ask,
\begin{enumerate}
    \item[] ``\textit{What is the probability of a patient being a smoker, given that he presented shortness of breath?}"
\end{enumerate}
We can extend this to a forensic context. Suppose a crime scene is investigated where a severely burned body is found. The forensic analyst can then ask,
\begin{enumerate}
    \item[]``\textit{The burn victim is found with a protruded tongue, was the victim exposed to fire before death or after?}"
\end{enumerate}
Consider now a financial service context where a young prospective home owner is declined a loan. The service provider can then ask,
\begin{enumerate}
    \item[]``\textit{Did the prospective owner not qualify for the home loan because of his age?}"
\end{enumerate}
From these examples, we see that explanation of reasoning is used where questions are asked in the context of single variable outcomes for diagnosis. 

\subsection{Evidence}
When we are interested in the subset of variables that describes specific scenarios, we use explanation of evidence methods. For example, in our Asia example the doctor may ask,
\begin{enumerate}
    \item[]``\textit{Which diseases are most probable to the symptoms presented by the patient?}"
\end{enumerate}
or
\begin{enumerate}
    \item[]``\textit{Which diseases are most relevant to the symptoms presented by the patient?}''
\end{enumerate}
In a forensic context, the forensic analyst investigating a crime scene may ask the following question,
\begin{enumerate}
    \item[]``\textit{What are the most relevant causes of death, given the victim is found with a severely burned body and protruded tongue?}"
\end{enumerate}
Similarly this can be applied to fraud detection. Suppose the analyst investigates the credit card transactions of a consumer. The analyst can then ask,  
\begin{enumerate}
    \item[]``\textit{What are the most probable transaction features that contributed to the flagging of this consumer?}"
\end{enumerate}
Explanation of evidence can also be used to provide explanations for financial service circumstances. For example, if a prospective home owner is turned down for a loan, he may ask the service provider which features in his risk profile are more relevant (contributed most) to being turned down.

\subsection{Decisions}
Explanation of decisions typically asks the following questions \textit{``Do we have enough evidence to make a decision?"}, and if not, \textit{``what additional evidence is required to make a decision?"}. For example, in our Asia example we can ask,
\begin{enumerate}
    \item[]``\textit{Do we have enough evidence on the symptoms presented to make a decision on the disease?}"
\end{enumerate}
or
\begin{enumerate}
    \item[]``\textit{Since we are not yet able to determine the disease,  what additional information -- test, underlying symptoms, comorbidities -- is required to make a decision?}"
\end{enumerate}
Applied to forensic investigations, this can be used to answer questions relating to crime scene investigations. The analyst may ask questions regarding the actual evidence collected from the crime scene, i.e., if enough evidence is collected to rule a crime as a homicide or what additional evidence is required to rule the crime as a homicide. Should they investigate further or is the evidence that is already collected enough to make an informed decision?

\section{Conclusion}\label{conclusion}
The development of AI systems has faced incredible advances in recent years. We are now exposed to these systems on a daily basis, such as product recommendation systems used by online retailers. However, these systems are also being implemented by medical practitioners, forensic analysts and financial services -- application areas where decisions directly influence the lives of humans. It is because of these high-risk application areas that progressively more interest is given to the explainability of these systems.

This paper addresses the problem of explainability in BNs. We first explored the state of explainable AI and in particular BNs, which serves as a foundation for our XBN framework. We then presented a taxonomy to categorise XBN methods in order to emphasise the benefits of each method given a specific usage of the BN model. This XBN taxonomy will serve as a guideline, which will enable end-users to understand how and why predictions were made and will, therefore, be able to better communicate how outcomes were obtained based on these predictions.

The XBN taxonomy consists of explanation of reasoning, evidence and decisions. Explanation of the model is reserved for future work, since the taxonomy described in this paper is focused on how and why predictions were made and not on the model-construction phase. Other future research endeavours include the addition of more dimensions and methods to the XBN taxonomy -- this involves more statistical-based methods and the incorporation of causability (which also addresses the quality of explanations) -- as well as applying this taxonomy to real-world applications.

%
%
%
\bibliographystyle{splncs04}
\bibliography{Paper.bib}

\end{document}